\documentclass[conference]{IEEEtran}
\IEEEoverridecommandlockouts

\usepackage{cite}
\usepackage{hyperref}
\usepackage{amsmath,amssymb,amsfonts}
\usepackage{multicol, multirow}
\usepackage{algorithmic}
\usepackage{graphicx}
\usepackage{textcomp}
\usepackage{xcolor}
\def\BibTeX{{\rm B\kern-.05em{\sc i\kern-.025em b}\kern-.08em
    T\kern-.1667em\lower.7ex\hbox{E}\kern-.125emX}}
\begin{document}

\title{REAct: Rational Exponential Activation for Better Learning and Generalization in PINNs}

\author{\IEEEauthorblockN{Sourav Mishra\IEEEauthorrefmark{1}\IEEEauthorrefmark{3}, Shreya Hallikeri\IEEEauthorrefmark{4}\IEEEauthorrefmark{5}\thanks{\IEEEauthorrefmark{5}work was done during her internship at the Artificial Intelligence and Robotics Lab, Department of Aerospace Engineering, IISc} and
Suresh Sundaram\IEEEauthorrefmark{2}\IEEEauthorrefmark{3}
\IEEEauthorblockA{\IEEEauthorrefmark{3}Department of Aerospace Engineering,
Indian Institute of Science (IISc), Bangalore;
\IEEEauthorrefmark{4}PES University, Bangalore\\
\texttt{Email: \{\IEEEauthorrefmark{1}souravmishr1, \ \IEEEauthorrefmark{2}vssuresh\}@iisc.ac.in, \IEEEauthorrefmark{4}shreya.hallikeri@gmail.com}}}}

\maketitle

\begin{abstract}
Physics-Informed Neural Networks (PINNs) offer a promising approach to simulating physical systems. Still, their application is limited by optimization challenges, mainly due to the lack of activation functions that generalize well across several physical systems. Existing activation functions often lack such flexibility and generalization power. To address this issue, we introduce Rational Exponential Activation (REAct), a generalized form of tanh consisting of four learnable shape parameters. Experiments show that REAct outperforms many standard and benchmark activations, achieving an MSE three orders of magnitude lower than tanh on heat problems and generalizing well to finer grids and points beyond the training domain. It also excels at function approximation tasks and improves noise rejection in inverse problems, leading to more accurate parameter estimates across varying noise levels. 
\end{abstract}


\section{Introduction and Background}
Physics-Informed Neural Networks (PINNs) have recently made significant progress in modeling physical systems by incorporating physical laws, expressed as ordinary and partial differential equations (ODEs and PDEs), into the training process as soft constraints. Traditional methods for simulating physical systems are either data-driven, which struggles with sparse or noisy data, or numerical, which are computationally expensive. PINNs offer a compromise between the two, being able to learn from sparse data unlike data-driven methods and more computationally efficient than numerical approaches \cite{raissi2019physics, markidis2021old}. As a result, PINNs have been successfully applied to model fluid flow \cite{cai2021physics-fluid}, heat flow \cite{cai2021physics-heat}, control system design \cite{furfaro2022physics, arnold2021state}, electromagnetics \cite{khan2022physics}, nano-optics and metamaterials \cite{nano_mater} and several other areas. However, despite their wide applicability, PINNs face optimization challenges due to the use of PDE-based loss functions, which can lead to ill-conditioned training \cite{failure_mode_pinn, gradient_pathology, NTK}. Activation functions are a key factor affecting PINN optimization as they determine how well the network captures the underlying dynamics of a physical system \cite{act_import}.

Activation functions play a key role in neural networks by introducing non-linearity, enabling models to capture complex patterns, making them an important research area \cite{ranjan2020understanding}. In continuous settings like those in PINNs, selecting activation functions becomes even more critical, as it impacts the network's ability to represent complex physical signals \cite{vortex, sinusoid_space, periodic}. The optimal activation function is often problem-specific; for instance, recent studies show that hyperbolic tangent can lead to instability in simulating certain dynamics \cite{vortex}, while sinusoidal functions offer smoother optimization \cite{sinusoid_space}. Therefore, the dynamics of the system being modeled plays a key role in choosing a suitable activation function for PINNs \cite{vortex, ABU}, unlike supervised learning problems where Relu is universally used regardless of the modality of the data \cite{relu, dying_relu}. Vibrating systems might exhibit resonance, leading to unusually high output magnitudes, making normalization techniques of little use in modeling the dynamics.




For PINNs to accurately model physical systems, their activation functions must meet crucial requirements: they need to be smooth and continuously differentiable to handle physics-informed loss functions, which involve higher-order derivatives. Additionally, the function should allow unbounded outputs, unlike tanh and sin, which are restricted between -1 and 1. Furthermore, activation functions must avoid saturation to prevent vanishing gradients, which hinder learning \cite{gradient_pathology}. As conventional activation functions used in classification tasks don’t satisfy the above criteria, recent PINN literature is taking a data-driven approach to design more generalizable activation functions \cite{jagtap2020adaptive, jagtap2020locally, STan}. These methods include learnable parameters in activation functions, improving convergence rate and solution quality \cite{jagtap2020adaptive, STan} by dynamically altering the loss landscape during training. The Adaptive Blending Unit (ABU) PINNs \cite{ABU} optimizes the search for the best activation function at each layer by learning a convex combination of a pre-selected set of standard activation functions, allowing the network to capture system-specific features.

Despite their benefits, the above activations offer limited control over key shape characteristics like zero crossings, frequency, saturation regions, and convexity, restricting their generalization capabilities. The performance of ABU-PINNs is further constrained by the learnable convex combination of a fixed set of activations chosen a-priori. Better performance comes at a higher computational cost by including more diverse activation functions in the set. Additionally, it may be desirable to have controlled saturation beyond some range to reject the influence of noise in estimating system parameters from noisy sensor data in inverse problems.
To overcome these issues, we introduce Rational Exponential Activation (REAct), a more generalized version of the hyperbolic tangent with four learnable parameters. REAct improves PINN performance on forward problems, reducing MSE by 3 orders of magnitude on the heat problem and generalizing well to finer grids and to points outside the training domain. It also captures the variability in complicated functions, leading to better function approximation accuracy, and demonstrates accurate parameter estimation in inverse problems across a range of noise levels.

\section{Method}

\subsection{Physics Informed Neural Networks (PINNs)}
Consider an Initial Boundary Value Problem (IBVP) described by a Partial Differential Equation (PDE) $\mathcal{N}(u, x, t) = 0$ over the domain $(x, t) \in [0, L] \times [0, T]$, with known initial conditions $u(x_i, t=0)$ at points $\{x_i\}_{i=1}^{N_I}$, and boundary conditions at $x=0$ and $x=L$ for time instants $\{t_b\}_{b=1}^{N_B}$. The PINN approximates the solution of the IBVP $u(x, t)$ as $\hat{u}(x, t)$. $u(x, t)$ may be known at the points $\{(x_d, t_d)\}_{d=1}^{N_D}$. To enforce the dynamics, the physics loss is computed as the PDE residual at collocation points $\{(x_c, t_c)\}_{c=1}^{N_C}$. The model outputs must also satisfy initial and boundary conditions while matching the given data. Each component has its respective loss given below:

\begin{align}
    \mathcal{L}_{phy} = \dfrac{1}{N_C}\sum_{c=1}^{N_C}(\mathcal{N}(\hat{u}_c, x_c, t_c))^2 \\
    \mathcal{L}_{IC} = \dfrac{1}{N_I}\sum_{i=1}^{N_I}(\hat{u}(x_i, t=0) - u(x_i, t=0))^2 \\
    \mathcal{L}_{BC} = \dfrac{1}{N_B}\sum_{b=1}^{N_B}(\hat{u}(x=0, t_b) - u(x=0, t_b))^2 \nonumber\\+ (\hat{u}(x=L, t_b) - u(x=L, t_b))^2 \\
    \mathcal{L}_{data} = \dfrac{1}{N_D}\sum_{d=1}^{N_D}(\hat{u}(x_d, t_d) - u(x_d, t_d))^2
    \label{eq: Physics, IC, BC, and data}
\end{align}

The total loss is a weighted sum of these components:

\begin{equation}
    \mathcal{L} = \lambda_p\mathcal{L}_{phy} + \lambda_I\mathcal{L}_{IC} + \lambda_B\mathcal{L}_{BC} + \lambda_d\mathcal{L}_{data}
    \label{eq: total loss}
\end{equation}

where $\lambda_p, \lambda_I, \lambda_B, \lambda_d$ are hyperparameters controlling the weight of each loss term.

\subsection{Rational Exponential Activation (REAct)}
Our motivation for designing more generalizable activation functions for PINNs is based on the following points:

\begin{itemize}
    \item They must be smooth and differentiable for calculating the physics loss.
    \item Outputs should be unbounded, as the magnitude of PINN outputs is unknown in advance.
    \item Control over key shape properties (e.g., zero crossings, frequency, convexity) to adapt to several signals and mitigate noisy measurements in inverse problems.
    \item More flexibility and fewer learnable parameters than ABU-PINN.
\end{itemize}

Common activations lack these features, particularly in controlling shape properties. Since tanh and sin are widely used in PINNs \cite{jagtap2020adaptive, sinusoid_space}, arise in the solution of many PDEs and ODEs, and are related to the exponential family, we introduce a generalized activation called Rational Exponential Activation (REAct) with four learnable shape parameters:

\begin{equation}
    REAct(x) = \dfrac{1 - \exp(ax + b)}{1 + \exp(cx + d)}
    \label{eq: REAct}
\end{equation}

\begin{figure}[t]
    \centering
    \includegraphics[width=0.9\linewidth]{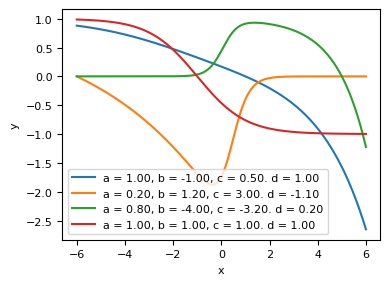}
    \caption{REAct for different values of shape parameters}
    \label{fig: Figure REAct}
\end{figure}

REAct meets these criteria and is in the same form as the activations in \cite{jagtap2020adaptive, STan}. Therefore, PINNs using REAct avoid suboptimal critical points, as proven in \cite{jagtap2020adaptive, STan}. Figure \ref{fig: Figure REAct} shows REAct for various values of its shape parameters, showing it is more flexible than STan \cite{STan}.

\section{Experiments and Results}

\subsection{Setup}
We evaluate REAct against various standard activation functions, including Relu \cite{relu}, sigmoid, tanh, sin, STan \cite{STan}, and ABU-PINN \cite{ABU}, across three tasks: forward simulations, function approximation, and inverse experiments. ABU-PINN combines Relu, sigmoid, sin, tanh, and softplus with softmax-normalized learnable weights, ensuring a convex linear combination. Five IBVPs are used for forward simulations, including an underdamped system with a damping ratio of 0.5 and a natural frequency of 3.0 rad/s. A 1D heat problem with Dirichlet boundary conditions and $u(x, t=0) = \sin(\pi x)$ is also considered. More details on the IBVPs can be found in \cite{lu2021deepxde}. Testing uses a finer grid for Allen Cahn and Underdamped vibration equations with ten times more points along each coordinate. For the other equations, the colocation points are uniformly spread in the problem domain, and only those in the testing domain are used for testing. Performance is assessed using the $L_2$ relative error, Mean Squared Error (MSE), Mean Absolute Error (MAE), and Explained Variance Score (EVS) \cite{kapoor2024neural}. All experiments are implemented with PyTorch \cite{PyTorch} and performed on an NVIDIA GeForce RTX 3090 Ti GPU with $\lambda_p, \lambda_I, \lambda_B, \lambda_d$ set to 1. Code is available at \texttt{\url{https://github.com/srvmishra/REAct}.}

\begin{table*}[t]
\centering
\caption{Experimental Settings for forward experiments}
\label{tab: Table settings}
\begin{tabular}{ccccccccc}
\hline
Equation                                                         & \begin{tabular}[c]{@{}c@{}}Train \\ domain\end{tabular}                  & \begin{tabular}[c]{@{}c@{}}Test \\ domain\end{tabular}                   & \begin{tabular}[c]{@{}c@{}}\# Space \\ points\end{tabular} & \begin{tabular}[c]{@{}c@{}}\# Time \\ points\end{tabular} & Optimizer & \begin{tabular}[c]{@{}c@{}}Learning\\ Rate\end{tabular} & Iterations & \begin{tabular}[c]{@{}c@{}}Model\\ Size\end{tabular} \\ \hline
Allen Cahn                                                       & \begin{tabular}[c]{@{}c@{}}$x$: {[}-1, 1{]}\\ $t$: {[}0, 1{]}\end{tabular}   & \begin{tabular}[c]{@{}c@{}}$x$: {[}-1, 1{]}\\ $t$: {[}0, 1{]}\end{tabular}   & 100                                                        & 100                                                       & RMSprop   & 1E-4 & 50000 & \begin{tabular}[c]{@{}c@{}}Input: 2, Output: 1\\ Hidden: 32 $\times$ 3\end{tabular}                                                   \\ \hline
Burgers                                                          & \begin{tabular}[c]{@{}c@{}}$x$: {[}-1, 1{]}\\ $t$: {[}0, 0.8{]}\end{tabular} & \begin{tabular}[c]{@{}c@{}}$x$: {[}-1, 1{]}\\ $t$: {[}0.8, 1{]}\end{tabular} & 256                                                        & 100                                                       & RMSprop   & 1E-4 & 20000 & \begin{tabular}[c]{@{}c@{}}Input: 2, Output: 1\\ Hidden: 32 $\times$ 3\end{tabular}                                                    \\ \hline
Diffusion                                                        & \begin{tabular}[c]{@{}c@{}}$x$: {[}-1, 1{]}\\ $t$: {[}0, 0.8{]}\end{tabular} & \begin{tabular}[c]{@{}c@{}}$x$: {[}-1, 1{]}\\ $t$: {[}0.8, 1{]}\end{tabular} & 100                                                        & 100                                                       & Adam      & 1E-3 & 30000 & \begin{tabular}[c]{@{}c@{}}Input: 2, Output: 1\\ Hidden: 30 $\times$ 6\end{tabular}                                                    \\ \hline
Heat                                                             & \begin{tabular}[c]{@{}c@{}}$x$: {[}0, 1{]}\\ $t$: {[}0, 0.8{]}\end{tabular}  & \begin{tabular}[c]{@{}c@{}}$x$: {[}0, 1{]}\\ $t$: {[}0.8, 1{]}\end{tabular}  & 100                                                        & 100                                                       & RMSprop   & 1E-4 & 50000 & \begin{tabular}[c]{@{}c@{}}Input: 2, Output: 1\\ Hidden: 48 $\times$ 3\end{tabular}                                                    \\ \hline
\begin{tabular}[c]{@{}c@{}}Under damped\\ Vibration\end{tabular} & $t$: {[}0, 1{]}                                                            & $t$: {[}0, 1{]}                                                            & -                                                       & 1000                                                     & Adam      & 1E-3 & 50000 & \begin{tabular}[c]{@{}c@{}}Input: 2, Output: 1\\ Hidden: 48 $\times$ 3\end{tabular}                                                    \\ \hline
\end{tabular}
\end{table*}

\subsection{Forward Experiments}
In forward experiments, a PINN is used to simulate the IBVP given its governing equation $\mathcal{N}(u, x, t) = 0$ and the initial and boundary conditions. Only $\mathcal{L}_{phy}$, $\mathcal{L}_{IC}$ and $\mathcal{L}_{BC}$ are used to train the PINN for forward problems. The problems are simulated using the settings mentioned in Table \ref{tab: Table settings}, and the results are given in Table \ref{tab: Table forward}. REAct gives the best results for all the forward problems, giving the maximum improvements on the heat problem - 0.008 decrease on MAE, and 3 orders of magnitude decrease on MSE compared to STan. Results also indicate that PINNs using REAct can generalize well beyond the training time interval (Burgers, Diffusion, and Heat Equations), and even to finer testing domains (Allen Cahn and Underdamped Vibration Equations).

\begin{table*}[]
\centering
\caption{Results of forward experiments. $\downarrow$ indicates lower values are better and vice versa.}
\label{tab: Table forward}
\begin{tabular}{clcccccccc}
\hline
Equation                                                                           & \multicolumn{1}{c}{Metric} & \multicolumn{1}{c}{ReLU} & \multicolumn{1}{c}{Sigmoid} & \multicolumn{1}{c}{tanh(x)} & \multicolumn{1}{c}{sin(x)} & \multicolumn{1}{c}{Softplus} & \multicolumn{1}{c}{STan} & \multicolumn{1}{c}{ABU} & \multicolumn{1}{c}{REAct} \\ \hline
\multirow{4}{*}{Allen Cahn}                                                        & L2 rel. ($\downarrow$)                    & 0.9922                   & 0.799                       & 0.8056                      & 0.983                      & 1.0539                      & 0.7576                   & 0.9592   & \textbf{0.6686}           \\
                                                                                   & MSE ($\downarrow$)                        & 0.4905                   & 0.3181                      & 0.3233                      & 0.4814                     & 0.5534                      & 0.286                    & 0.4585   & \textbf{0.2228}           \\
                                                                                   & MAE ($\downarrow$)                        & 0.4567                   & 0.3031                      & 0.306                       & 0.3896                     & 0.4903                      & 0.2852                   & 0.5601  & \textbf{0.2505}           \\
                                                                                   & EVS ($\uparrow$)                        & 0.1161                   & 0.4866                      & 0.4792                      & 0.2575                     & 0.0496                      & 0.5382                   & 0.0691  & \textbf{0.6372}           \\ \hline
\multirow{4}{*}{Burgers}                                                           & L2 rel. ($\downarrow$)                    & 0.6675                   & 0.5958                      & 0.2043                      & 0.237                      & 0.632                       & 0.1509                   & 0.4532     & \textbf{0.1496}           \\
                                                                                   & MSE ($\downarrow$)                        & 0.0934                   & 0.0744                      & 0.0087                      & 0.0118                     & 0.0837                      & 0.0048                   & 0.043     & \textbf{4.7E-03}          \\
                                                                                   & MAE ($\downarrow$)                        & 0.1913                   & 0.1764                      & 0.0392                      & 0.0456                     & 0.1991                      & 0.0313                   & 0.1032  & \textbf{0.0298}           \\
                                                                                   & EVS ($\uparrow$)                        & 0.5545                   & 0.645                       & 0.9586                      & 0.944                      & 0.6023                      & 0.9777                   & 0.7961   & \textbf{0.9805}           \\ \hline
\multirow{4}{*}{Diffusion}                                                         & L2 rel. ($\downarrow$)                    & 0.9534                   & 0.0142                      & 0.0204                      & 0.0196                     & 0.0049                      & 0.0066                   & 0.1982    & \textbf{0.004}            \\
                                                                                   & MSE ($\downarrow$)                        & 0.0749                   & 1.65E-05                    & 3.44E-05                    & 3.18E-05                   & 1.99E-06                    & 3.64E-06                 & 0.0032   & \textbf{1.32E-06}         \\
                                                                                   & MAE ($\downarrow$)                        & 0.2419                   & 0.003                       & 0.0043                      & 0.0037                     & 0.001                       & 0.0014                   & 0.0473  & \textbf{0.0008}           \\
                                                                                   & EVS ($\uparrow$)                        & 0.0911                   & 0.9998                      & 0.9996                      & 0.9996                     & \textbf{1}                  & \textbf{1}               & 0.9879    & \textbf{1}                \\ \hline
\multirow{4}{*}{Heat}                                                              & L2 rel. ($\downarrow$)                    & 32.3597                  & 0.921                       & 0.3879                      & 1.0113                     & 0.4089                      & 0.4214                   & 15.5288      & \textbf{0.0214}           \\
                                                                                   & MSE ($\downarrow$)                        & 0.4756                   & 0.0004                      & 0.0001                      & 0.0005                     & 0.0001                      & 0.0001                   & 0.1095  & \textbf{2.09E-07}         \\
                                                                                   & MAE ($\downarrow$)                        & 0.6207                   & 0.0151                      & 0.0079                      & 0.0211                     & 0.0084                      & 0.0088                   & 0.3014   & \textbf{0.0004}           \\
                                                                                   & EVS ($\uparrow$)                        & 0.351                  & 0.8425                       & 0.9463                      & 0.8187                     & 0.9567                      & 0.9786                   & 0.6493    & \textbf{0.9995}           \\ \hline
\multirow{4}{*}{\begin{tabular}[c]{@{}c@{}}Under damped \\ Vibration\end{tabular}} & L2 rel. ($\downarrow$)                    & 0.9097                   & 0.0011                      & 0.0013                      & 7.85E-05                   & 0.0035                      & 1.75E-06                 & 0.0049   & \textbf{1.52E-06}         \\
                                                                                   & MSE ($\downarrow$)                        & 0.2646                   & 3.71E-07                    & 5.10E-07                    & 1.97E-09                   & 7.59E-06                    & 9.77E-13                 & 7.59E-06        & \textbf{7.34E-13}         \\
                                                                                   & MAE ($\downarrow$)                        & 0.4047                   & 0.0006                      & 0.0007                      & 3.76E-05                   & 0.002                       & 7.92E-07                 & 0.0022   & \textbf{7.29E-07}         \\
                                                                                   & EVS ($\uparrow$)                        & 0.1419                   & \textbf{1}                  & \textbf{1}                  & \textbf{1}                 & \textbf{1}                  & \textbf{1}               & 0.9999    & \textbf{1}                \\ \hline
\end{tabular}
\end{table*}

\subsection{Function Approximations}
We consider the following functions for the function approximation tasks.

\begin{equation}
    f_1(x) = x^2\sin(2x), \ x \in [-\pi, \pi]
    \label{eq: Equation Parabolic Sinusoid}
\end{equation}

\begin{equation}
    f_2(x) = \frac{x^3 - x}{7}\sin(7x) + \sin(12x), \ x \in [-\pi, \pi]
    \label{eq: Equation Polynomial Sinusoid}
\end{equation}

\begin{equation}
    f_3(x) = \sin(2x + \pi/3)\sin(4x + \pi/6), \ x \in [0, 2\pi]
    \label{eq: Equation Sinusoidal Beats}
\end{equation}

In each case, 1000 points are uniformly sampled across the domain. Since this is a regression problem, the network is trained by minimizing $\mathcal{L}_{data}$ at these points only. Table \ref{tab: Table func. approx.} presents the results for function approximation tasks. REAct outperforms other activation functions, particularly excelling with $f_2(x)$ (Eq. \ref{eq: Equation Polynomial Sinusoid}). This indicates that REAct’s enhanced flexibility, arising from its learnable shape parameters, allows it to capture variations in polynomial functions and sinusoids of various frequencies more effectively than other activation functions. Conversely, ABU performs poorly due to its limited expressive power, as it lacks shape parameters within individual activations and relies only on a convex combination of pre-selected activations.

\begin{table*}
\centering
\caption{Results of function approximation tasks. $\downarrow$ indicates lower values are better and vice versa.}
\label{tab: Table func. approx.}
\begin{tabular}{clcccccccc}
\hline
Function                                                                        & Metric  & ReLU       & Sigmoid    & tanh(x) & Softplus  & sin(x)     & STan       & ABU    & REAct    \\ \hline
\multirow{4}{*}{\begin{tabular}[c]{@{}c@{}} $f_1(x)$ \\ Eq. \ref{eq: Equation Parabolic Sinusoid}\end{tabular}}  & L2 rel. ($\downarrow$) & 0.0137     & 0.0067     & 0.0224  & 0.0561     & 0.0116     & 0.0066     & 0.0086 & \textbf{0.0051}   \\
                                                                                & MSE ($\downarrow$)     & 0.0016     & 0.0004     & 0.0043  & 0.0269     & 0.0012     & 0.0004     & 0.0006  & \textbf{0.0002}   \\
                                                                                & MAE ($\downarrow$)     & 0.0372     & 0.0125     & 0.0293  & 0.1442     & 0.0274     & 0.0143     & 0.0157   & \textbf{0.0109}   \\
                                                                                & EVS ($\uparrow$)     & \textbf{1} & \textbf{1} & 0.9995  & 0.9987     & 0.9999     & \textbf{1} & 0.9999    & \textbf{1} \\ \hline
\multirow{4}{*}{\begin{tabular}[c]{@{}c@{}}$f_2(x)$ \\ Eq. \ref{eq: Equation Polynomial Sinusoid}\end{tabular}} & L2 rel. ($\downarrow$) & 0.1363     & 0.7392     & 0.0075  & 0.0125     & 0.0065     & 0.0107     & 0.007  & \textbf{0.0029}   \\
                                                                                & MSE ($\downarrow$)     & 0.0281     & 0.8257     & 0.0001  & 0.0002     & 0.0001     & 0.0002     & 0.0001   & \textbf{1.29E-05} \\
                                                                                & MAE ($\downarrow$)     & 0.0674     & 0.6014     & 0.0072  & 0.0118     & 0.0058     & 0.0106     & 0.0062   & \textbf{0.0026}   \\
                                                                                & EVS ($\uparrow$)     & 0.9812     & 0.4418     & 0.9999  & 0.9998     & \textbf{1} & 0.9999     & \textbf{1} & \textbf{1} \\ \hline
\multirow{4}{*}{\begin{tabular}[c]{@{}c@{}}$f_3(x)$ \\ Eq. \ref{eq: Equation Sinusoidal Beats}\end{tabular}}    & L2 rel. ($\downarrow$) & 0.3372     & 0.473      & 0.571   & 0.3513     & 0.0278     & 0.033      & 0.4166   & \textbf{0.0239}   \\
                                                                                & MSE ($\downarrow$)     & 0.0284     & 0.0559     & 0.0815  & 0.0308     & 0.0002     & 0.0003     & 0.0434 & \textbf{0.0001}   \\
                                                                                & MAE ($\downarrow$)     & 0.101      & 0.1223     & 0.2213  & 0.1243     & 0.011      & 0.0136     & 0.1473   & \textbf{0.0092}   \\
                                                                                & EVS ($\uparrow$)     & 0.8879     & 0.7763     & 0.6774  & 0.8927     & 0.9996     & 0.9996     & 0.8292   & \textbf{0.9997}   \\ \hline
\end{tabular}
\end{table*}

\subsection{Inverse Experiments}
Inverse experiments are performed to estimate unknown parameters in the governing equations or initial/boundary conditions from possibly noisy sensor measurements. We consider two inverse problems: a 1D inverse heat problem to determine thermal diffusivity ($\alpha$) and a 1D inverse wave problem to find wave velocity ($c$). The heat problem uses the same conditions as the forward experiments, with the wave problem having Dirichlet boundary conditions at $x = 0$ and $x = 2$, and an initial displacement profile $u(x, 0) = \sin(\pi x/2)$. A wave velocity of $c = 2$ m/s is assumed. For the heat problem, a thermal diffusivity $\alpha = 0.4$ $\text{m}^2/\text{s}$ is assumed. For both problems, 10,000 points are uniformly sampled to apply $\mathcal{L}_{phy}$, $\mathcal{L}_{IC}$, and $\mathcal{L}_{BC}$. Gaussian noise sampled from $\mathcal{N}(0, 0.1)$ is added to the analytical solution values at 5,000 of these points, and this noisy data is used to impose $\mathcal{L}_{data}$. $\alpha$ and $c$ are intialized using the uniform distribution $U[0.2, 2.5]$ (see Table \ref{tab: Table inverse}). The network and problem parameters are jointly optimized using the Adam optimizer \cite{adam} (learning rate 0.001, 50000 iterations) for the heat problem and the RMSprop optimizer \cite{rmsprop} (learning rate 0.0001, 75000 iterations) for the wave problem. The final estimates of $\alpha$ and $c$ and percentage errors in estimation are reported. Table \ref{tab: Table inverse} shows that REAct provides the best estimate for wave velocity (1.9968) with the lowest percentage error (0.16\%) and performs well in the heat problem, alongside tanh, with an estimate of 0.3999 for $\alpha$ and a percentage error of 0.025\%. 

\begin{table*}[]
\centering
\caption{Results of inverse experiments. $\downarrow$ indicates lower values are better}
\label{tab: Table inverse}
\begin{tabular}{clccccccccc}
\hline
\multicolumn{1}{c}{Problem}                                                    & Metric   & \multicolumn{1}{c}{initial value} & \multicolumn{1}{c}{ReLU} & \multicolumn{1}{c}{Sigmoid} & \multicolumn{1}{c}{tanh(x)} & \multicolumn{1}{c}{sin(x)} & \multicolumn{1}{c}{Softplus} & \multicolumn{1}{c}{STan} & \multicolumn{1}{c}{ABU} & \multicolumn{1}{c}{REAct} \\ \hline
\multirow{2}{*}{\begin{tabular}[c]{@{}l@{}}Heat \\ $\alpha = 0.3$\end{tabular}} & estimate & 1.7081 & 1.7081                   & 0.3997                      & \textbf{0.3999}             & 0.3993                     & 0.3981                     & 0.4006                   & 1.5335                   & \textbf{0.3999}           \\
                                                                                & \% error ($\downarrow$) & 327.025 & 327.025                  & 0.075                       & \textbf{0.025}              & 0.175                      & 0.475                      & 0.15                     & 283.375                  & \textbf{0.025}            \\ \hline
\multirow{2}{*}{\begin{tabular}[c]{@{}l@{}}Wave\\ $c = 2.0$\end{tabular}}       & estimate & 1.4623 & 1.4623                   & 1.9871                      & 1.9962                      & 1.9951                     & 1.9814                     & 1.9942                   & 1.7735                   & \textbf{1.9968}           \\
                                                                                & \% error ($\downarrow$) & 26.885 & 26.885                   & 0.645                       & 0.19                        & 0.245                      & 0.93                       & 0.29                     & 11.325                   & \textbf{0.16}             \\ \hline
\end{tabular}
\end{table*}

\subsection{Ablations}
Ablation studies on inverse problems are carried out to test the noise rejecting capabilities of REAct using the earlier settings. However, the level of noise added to the data points is progressively increased by varying the standard deviation of the noise distribution. Table \ref{tab: Table Inverse Ablations} shows the percentage errors in the estimated thermal diffusivity ($\alpha$) and wave velocity ($c$). Differences in initial parameter values between Tables \ref{tab: Table inverse} and \ref{tab: Table Inverse Ablations} account for variations in percentage error, even with the same noise standard deviation of 0.1. REAct outperforms STan and ABU across a range of noise levels, particularly at higher standard deviation values, showing it is more effective at mitigating noise. This observation suggests that REAct can adapt its saturation regions to filter out noise from high output ranges, making it more suitable for parameter estimation from noisy data than non-saturating functions like ABU and STan.

\begin{table}[]
\centering
\caption{Ablation Studies on Inverse Problems at Various Noise Levels. Percentage Error Values are reported.}
\label{tab: Table Inverse Ablations}
\begin{tabular}{lcccc}
\hline
Problem                                                                                       & noise std & STan   & ABU     & REAct  \\ \hline
\multirow{4}{*}{\begin{tabular}[c]{@{}l@{}}Heat \\ initial \\ $\alpha = 1.8116$\end{tabular}} & 0.1       & 0.1205 & 271.502 & \textbf{0.0285} \\
                                                                                              & 0.5       & 0.2811 & 182.224 & \textbf{0.0309} \\
                                                                                              & 1         & \textbf{0.0769} & 77.0141 & 0.1305 \\
                                                                                              & 5         & 3.2664 & 214.565 & \textbf{0.6878} \\ \hline
\multirow{4}{*}{\begin{tabular}[c]{@{}l@{}}Wave \\ initial \\ $c = 1.4623$\end{tabular}}      & 0.1       & 0.29 & 11.325 & \textbf{0.16} \\
                                                                                              & 0.5       & 0.3419 & 3.8189  & \textbf{0.1478} \\
                                                                                              & 1         & 0.5117 & 9.9029  & \textbf{0.0939} \\
                                                                                              & 3         & 0.8284 & 6.7969  & \textbf{0.8093} \\ \hline
\end{tabular}
\end{table}

\section{Conclusions}
Selection of proper activation functions for PINNs requires prior knowledge of the dynamics of the phenomena being modeled, making them highly problem-specific. It is one of the main contributors to optimization issues in PINNs. Current literature seeks to develop activation functions for PINNs that can generalize well across diverse physical systems. This work proposes REAct, a novel, more generalized, and flexible version of the hyperbolic tangent activation function with four learnable shape parameters for better optimization and generalization of PINNs. REAct outperforms standard activation functions and recent benchmark activations (STan and ABU) on five well-known forward problems, obtaining three orders of magnitude lower MSE on the heat problem than STan. It generalizes well on finer grids as well as on points beyond the training domain. Its effectiveness on function approximation problems is shown by its improved ability to capture variations in complicated functions and its enhanced noise rejection ability makes it better suited for inverse problems. 








\bibliographystyle{IEEEbib.bst}
\bibliography{refs.bib}

\end{document}